\documentclass{article}
\usepackage{arxiv}
\usepackage[utf8]{inputenc} 
\usepackage[T1]{fontenc}    
\usepackage{hyperref}       
\usepackage{amsmath,amssymb,amsfonts}
\usepackage{url}            
\usepackage{booktabs}       
\usepackage{amsfonts}       
\usepackage{nicefrac}       
\usepackage{microtype}      
\usepackage{cleveref}       
\usepackage{lipsum}         
\usepackage{graphicx}
\usepackage{xcolor}
\usepackage{appendix}
\usepackage{subcaption}
\usepackage{multicol}
\usepackage{doi}
\usepackage{comment}
\usepackage[ruled,vlined]{algorithm2e}
\usepackage{xcolor}

\SetCommentSty{mycommfont}
\usepackage[backend=biber, isbn=false, url=false, sorting=none]{biblatex}
\title{Choose Your Explanation: A Comparison of SHAP and Grad-CAM in Human Activity Recognition}

\date{\today}

\newif\ifuniqueAffiliation
\uniqueAffiliationtrue 

\ifuniqueAffiliation
\author{%
    \href{https://orcid.org/0009-0005-6310-408X}{\includegraphics[scale=0.06]{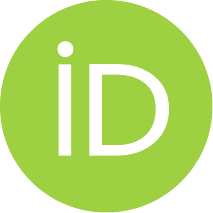}\hspace{1mm}Felix Tempel}$^{1}$,
    \href{https://orcid.org/0000-0003-0676-2324}{\includegraphics[scale=0.06]{orcid.pdf}\hspace{1mm}Daniel Groos}$^{2}$,
    \href{https://orcid.org/0000-0002-2469-1809}{\includegraphics[scale=0.06]{orcid.pdf}\hspace{1mm}Espen Alexander F. Ihlen}$^{2}$,
    \href{https://orcid.org/0000-0001-5532-0034}{\includegraphics[scale=0.06]{orcid.pdf}\hspace{1mm}Lars Adde}$^{3,4}$,
    \href{https://orcid.org/0000-0003-1820-6544}{\includegraphics[scale=0.06]{orcid.pdf}\hspace{1mm}Inga Strümke}$^{1}$ 
}

\date{
    $^1$Faculty of Informatics, Norwegian University of Science and Technology, Trondheim, Norway\\
    $^2$Faculty of Medicine and Health Sciences, Norwegian University of Science and Technology, Trondheim, Norway\\
    $^3$Department of Clinical and Molecular Medicine, NTNU, Trondheim, Norway\\
    $^4$Clinic of Rehabilitation, St. Olavs Hospital, Trondheim University Hospital, Norway\\
    \texttt{felix.e.f.tempel@ntnu.no, daniel.groos@ntnu.no, espen.ihlen@ntnu.no}\\
    \texttt{lars.adde@ntnu.no, inga.strumke@ntnu.no}
}
\fi

\renewcommand{\headeright}{}

\usepackage[colorinlistoftodos,prependcaption,textsize=tiny]{todonotes}

\hypersetup{
pdfauthor={Felix Tempel},
}

\begin{document}
\maketitle

\begin{abstract}


Explaining machine learning (ML) models using eXplainable AI (XAI) techniques has become essential to make them more transparent and trustworthy.
This is especially important in high-risk environments like healthcare, where understanding model decisions is critical to ensure ethical, sound, and trustworthy outcome predictions.
However, users are often confused about which explanability method to choose for their specific use case.
We present a comparative analysis of two explainability methods, Shapley Additive Explanations (SHAP) and Gradient-weighted Class Activation Mapping (Grad-CAM), within the domain of human activity recognition (HAR) utilizing graph convolutional networks (GCNs).
By evaluating these methods on skeleton-based input representation from two real-world datasets, including a healthcare-critical cerebral palsy (CP) case, this study provides vital insights into both approaches' strengths, limitations, and differences, offering a roadmap for selecting the most appropriate explanation method based on specific models and applications. 
We quantitatively and quantitatively compare the two methods, focusing on feature importance ranking and model sensitivity through perturbation experiments. 
While SHAP provides detailed input feature attribution, Grad-CAM delivers faster, spatially oriented explanations, making both methods complementary depending on the application's requirements. 
Given the importance of XAI in enhancing trust and transparency in ML models, particularly in sensitive environments like healthcare, our research demonstrates how SHAP and Grad-CAM could complement each other to provide model explanations.

\end{abstract}


\section{Introduction}
\label{sec:introduction}
Significant progress has been made in the development of Graph Convolution Networks (GCNs) for Human Action Recognition (HAR) in recent years.
GCNs are considered a practical solution to classify human movement and activities based on skeleton data compared to approaches using raw video data as the input.
This is due to the fact that raw video data is often polluted with noise, occluded body parts, and background information that is not informative for action recognition or movement classification.
Furthermore, a skeleton representation is free of skin and body composition bias since this information is not embedded in the skeleton sequence.
This makes GCNs a practical choice within the HAR domain if the skeleton representation is available.
A considerable proportion of the research is dedicated to improving the performance of GCN architectures on widely used benchmark datasets, including NTU RGB+D 60/120 \cite{liuNTURGB+D1202020, shahroudyNTURGB+DLarge2016} and Kinetics \cite{kayKineticsHumanAction2017}.
However, less attention has been paid to understanding and explaining the internal workings of the developed architectures. 
Concurrently, these models' increasing complexity and size have reduced the explainability aspect, turning them into ``black boxes''.
Still, understanding the inner workings of such systems is crucial, particularly when they are used in high-risk and sensitive environments such as healthcare, where the outcomes of model decisions can have significant ethical and practical implications \cite{bharatiReviewExplainableArtificial2024, chaddadSurveyExplainableAI2023, lohApplicationExplainableArtificial2022, sadeghiReviewExplainableArtificial2024}.
In such contexts, the explainability of models becomes not only a technical concern but also a matter of trust, accountability, and safety \cite{tjoaSurveyExplainableArtificial2021}.

The need for explainability in machine learning (ML) models is pushing the adoption of eXplainable Artificial Intelligence (XAI) techniques to describe the inner workings of GCNs \cite{montavonMethodsInterpretingUnderstanding2018}.
XAI provides tools for understanding how ML models generate particular predictions to facilitate trust and confidence among end-users and developers. 
This is done by translating the abstract and high-dimensional concepts underlying these models into a space accessible and comprehensible for humans.
XAI is divided into two key categories: interpretability and explainability \cite{vishwarupeExplainableAIInterpretable2022}. 
Interpretability refers to establishing clear, transparent rules that define how an ML model arrives at its decisions.
These insights can then be used for adequate clinical adoption and to reduce the risks emerging from biases in the ML models since physicians can understand the decision process. 
On the other hand, explainability entails creating a framework that explains the ML model's workings, often by representing its decision-making process in a simplified manner, making it accessible for humans to understand \cite{sadeghiReviewExplainableArtificial2024}.

In domains like healthcare, where decisions can directly impact patient outcomes, the explainability of ML models used in this field is critical \cite{koricaExplainableArtificialIntelligence2021, allgaierHowDoesModel2023, nazirSurveyExplainableArtificial2023}. 
Clinicians require assurance that ML models produce decisions based on relevant, reliable, and ethically sound data patterns \cite{a.SystematicReviewExplainable2023}. 
The lack of explainability of such systems could lead to dangerous scenarios, including the perpetuation of biases or the incorrect categorization of patient conditions, which might have significant clinical consequences \cite{gerlingsExplainableAIExplainable2022, kellyKeyChallengesDelivering2019}. 
Therefore, providing explainability by ``opening the black box'' of GCN-based HAR models is essential for their safe deployment and acceptance in sensitive environments such as healthcare \cite{vellidoImportanceInterpretabilityVisualization2020}. 

Despite the increasing relevance of XAI, its application within the HAR domain remains largely underexplored. 
To the extent that XAI methods have been studied for HAR, these focus primarily on three methods: Shapley Additive Explanations (SHAP) \cite{gaoAutomatingGeneralMovements2023, tempelExplainingHumanActivity2025}, Class Activation Mapping (CAM) \cite{songStrongerFasterMore2020} and its rectification Gradient-weighted Class Activation Mapping (Grad-CAM) \cite{dasGradientWeightedClassActivation2022}.
Beyond this, there remains a noticeable gap in the literature regarding the systematic comparison of these methods in the HAR field. 
Specifically, no work has been done to assess which explainability approach is better suited for different types of HAR models and datasets.
Further, there is a limited understanding of how different explainability methods vary in their outputs.
This gap presents a significant challenge, as the effectiveness and explainability of XAI methods can vary significantly and depend on the context and model architecture in which they are used.
For example, SHAP provides insights into each feature's importance but may fail to capture the spatial and temporal dynamics in the data.
Grad-CAM can highlight important regions for the model but may fall short of explaining the contribution of the individual input features. 
The lack of a comprehensive comparison of these methods with different datasets and HAR model types poses a challenge in identifying the most appropriate methods for model explanation.
Furthermore, clinicians and researchers should be provided with a quantitative comparison of these two XAI methods to assess which is best suited for their individual HAR model, dataset, and use case.
Understanding what to expect from each XAI method regarding explanation quality, transparency, and adaptability is essential to making informed decisions.
Hence, a systematic comparison of these two XAI methods is needed.
This can pave the way for trustworthy HAR models that can be safely and effectively implemented in real-world applications.





\subsection{Contributions}

\begin{itemize}
    \item Expanding an existing XAI framework to provide Grad-CAM explanations.
    \item First work to compare two explanation methods, SHAP and Grad-CAM, within the HAR domain, focusing on feature ranking, quantitative evaluation of the explanation's reliability, and a qualitative comparison.
    \item A nuanced understanding of the strengths and weaknesses of the two explanation approaches for GCNs within the HAR domain.
\end{itemize}

\section{Related Work}

\subsection{XAI Methods}
To provide explanations of ML models, a range of XAI methods have been applied \cite{bharatiReviewExplainableArtificial2024, chaddadSurveyExplainableAI2023, tjoaSurveyExplainableArtificial2021, allgaierHowDoesModel2023}. 
Among the most commonly used ones within the HAR domain are SHAP and Grad-CAM \cite{lundbergUnifiedApproachInterpreting2017, selvarajuGradCAMVisualExplanations2020}, which we apply in this study and discuss in greater detail in Sec.~\ref{sec:gradcam} and \ref{sec:shap}.
Other approaches within XAI include saliency maps \cite{simonyanDeepConvolutionalNetworks2014} or Local Interpretable Model-agnostic Explanations (LIME) \cite{ribeiroWhyShouldTrust2016}.

Saliency maps highlight the regions of an input that most influence a model's decision \cite{simonyanDeepConvolutionalNetworks2014}. 
These visualizations help identify key input features and can guide dimensionality reduction, as shown in \cite{yanResNetLikeCNNArchitecture2022}. 
While the use of saliency maps has been well validated in computer vision tasks \cite{simonyanDeepConvolutionalNetworks2014, adebayoSanityChecksSaliency2018}, direct applications in skeleton-based HAR are sparse, to the best of our knowledge.
Furthermore, saliency maps often provide a broad, unstable view of the model's decision-making process \cite{adebayoSanityChecksSaliency2018}.

LIME uses interpretable surrogate models to explain predictions  \cite{ribeiroWhyShouldTrust2016}. 
It does this by generating small input variations and training a simplified model to approximate how the complete model behaves locally \cite{ribeiroWhyShouldTrust2016}. 
While LIME has proven effective across various domains, its application in HAR remains unexplored, to the best of our knowledge. 
The challenge in applying LIME within HAR lies in generating kinematic sound perturbations for skeleton-based data since random changes to the body coordinates can easily create unrealistic human movements, making the explanations unreliable.

In summary, while various XAI methods offer promising tools for interpreting ML models, their adaptation to skeleton-based HAR remains an open research challenge, especially when preserving kinematic sound data is necessary.

\subsection{XAI in Human Activity Recognition}
In the domain of skeleton-based HAR, most XAI techniques rely on saliency-based methods such as CAM \cite{songStrongerFasterMore2020} or Grad-CAM \cite{dasGradientWeightedClassActivation2022, Song2022ConstructingSA}. 
Meanwhile, SHAP, which is widely used in a wide range of ML applications, has received less attention in HAR and was only applied in \cite{gaoAutomatingGeneralMovements2023, tempelExplainingHumanActivity2025}, to the best of our knowledge.

The first work using XAI within HAR is the work from \cite{songStrongerFasterMore2020}.
The authors use CAM to visualize important body key points on the two action classes, \textit{throwing} and \textit{drinking water} from the NTU RGB+D dataset \cite{shahroudyNTURGB+DLarge2016}.
However, their investigation is restricted to only a qualitative visual representation of the activated body key points for those two action classes.
This limits their applicability and quantitative comparability with other techniques.
The same authors expand this approach on a refined architecture in \cite{Song2022ConstructingSA}, where CAM is used again to produce visual explanations. 
However, these explanations also lack objective evaluation metrics, making comparing their effectiveness with other approaches difficult.
Also, a comparison of the activation maps to the earlier work \cite{songStrongerFasterMore2020} is missing.
It can be stated that the CAM approach does not address the contribution of individual input features, an area where SHAP could offer more profound insights.
This may be particularly interesting for the architecture used in \cite{Song2022ConstructingSA}, where the authors use multiple input branches to process features individually before fusing them into a common stream in the model.

In \cite{gaoAutomatingGeneralMovements2023}, SHAP is applied in a medical context for early cerebral palsy (CP) screening.
The model combines video data and secondary features like birth weight, sex, and gestational age to detect fidgety movements, indicating an increased risk for neurological dysfunction if they are absent or sporadic \cite{einspielerFidgetyMovementsTiny2016}. 
These features are then classified and combined into a normal and a risk probability for CP. 
However, the SHAP explanation is limited to the secondary input features, missing the opportunity to explain the primary input features (i.e., the video data).
SHAP would also have to be applied to the primary features to adequately explain the model, although this represents a more significant technical challenge.

Another work using SHAP within the medical HAR domain is \cite{tempelExplainingHumanActivity2025}, which this work builds upon.
Here, the authors use SHAP to explain the contributions of the primary input features on the model output utilizing two real-world datasets.
While this work is the first to apply SHAP on the primary input features, the direct comparison with the often-used Grad-CAM method within HAR is lacking.
A comparison between SHAP and Grad-CAM can provide a more complete evaluation of the individual strengths and weaknesses, which is not possible if only one explanation method is used.
This opens up the question of how methods like SHAP and Grad-CAM might complement or outperform each other in different HAR scenarios and datasets.

\section{Method}
\begin{figure}
    \centering
    \includegraphics[scale=1.1]{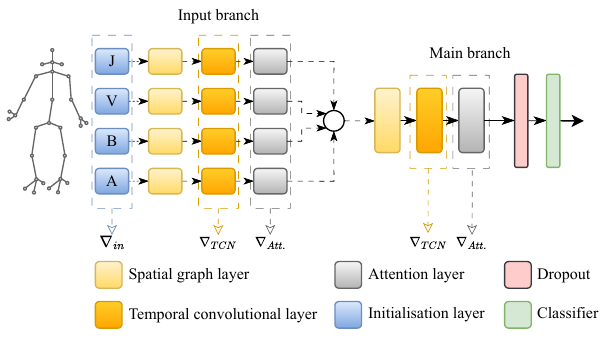}
    \caption{Model architecture with the locations where the gradients are obtained.
    The model consists of four input branches ($J, V, B, A$) processing different feature divisions obtained from the skeleton video.
    These features are fused later into a common main branch followed by the classifier.
    Our reference gradients  \(\nabla\) for the experiments are extracted after the attention activation (Att.) and after the temporal convolutional layer (TCN) in the main branch.}
    \label{fig:pipeline}
\end{figure}

This section outlines the two explainability techniques, SHAP and Grad-CAM, which are compared in this study.
Furthermore, the model architecture used to predict the skeleton data is briefly described, including at which layer we obtain the gradients for Grad-CAM; see a visualization in Fig.~\ref{fig:pipeline}. 
Finally, the perturbation technique used to evaluate the model's sensitivity is explained, including how specific parts of the architecture are perturbed based on the feature ranking from SHAP and Grad-CAM.

\subsection{Grad-CAM}\label{sec:gradcam}
Grad-CAM generates an explanation using the gradients from the final convolutional layer of an ML model \cite{selvarajuGradCAMVisualExplanations2020}. 
This is done by computing the gradients for class $c$ with respect to the activations of the feature maps in the chosen convolutional layer. 
Let $y^c$ represent the model's activation for class $c$ before the chosen target layer, corresponding to the model's prediction for class $c$.
The gradients of $y^c$ are calculated concerning each feature map activation $A^k$, where $A^k$ denotes the $k$-th feature map of the final convolutional layer. 
These gradients indicate how much each spatial location in the feature map contributes to the class prediction. 
Next, global average pooling is applied to the gradients, yielding the neuron importance weights
\begin{equation}
    \alpha_k^c = \frac{1}{Z} \sum_{i} \sum_{j} \frac{\partial y^c}{\partial A_{i,j}^k} \,,
\end{equation}

where $Z$ represents the number of spatial nodes in the feature map $A^k$, and $(i, j)$ are the spatial indices of the feature map. 
These importance weights $\alpha_k^c$ reflect the contribution of each feature map to the class prediction.

The final localization map, which highlights the relevant regions in the skeleton, is then computed as
\begin{equation}
\label{eq:grad}
    L_{\text{Grad-CAM}}^c = \text{ReLU} \left( \sum_{k} \alpha_k^c A^k \right) \,,
\end{equation}
where the ReLU activation ensures that only positive contributions (i.e., regions that are positively associated with class $c$) are retained.

We additionally apply the counterfactual explanation, where the gradients are negated \cite{selvarajuGradCAMVisualExplanations2020}.
This modification allows us to highlight the areas that decrease the class score, identifying the features that are either less important or negatively contribute to the prediction, i.e., \ contradict.
This way, we get the negated importance weights via
\begin{equation}
    \alpha_k^c = \frac{1}{Z} \sum_{i} \sum_{j} - \frac{\partial y^c}{\partial A_{i,j}^k} \,,
\end{equation}
and then compute the localization map with eq. \eqref{eq:grad}.

For skeleton-based HAR, Grad-CAM can be adapted to focus on the relevant spatial-temporal features within the skeletal data. 
In this context, the skeleton is typically represented as a graph, where nodes correspond to body key points, and edges represent their spatial relations. 
In GCNs, the final convolutional layer captures the spatial dependencies across the body key points, while the temporal information is encoded across different frames.

In \cite{dasGradientWeightedClassActivation2022}, Grad-CAM is extended to skeleton-based HAR tasks by adapting the method to handle the graph-structured data. 
Here, the gradients are computed for the graph convolutions, and the localization map indicates which body key points and frames are most influential in the model's prediction for a specific human activity. 
This approach can be used to visualize the contributions of different joints and frames, facilitating an understanding of which parts of the body and movement sequences are most relevant for a given activity at a specific time.

\subsection{SHAP}\label{sec:shap}
SHAP is a framework for explaining ML model predictions by attributing an importance score to each feature \cite{lundbergUnifiedApproachInterpreting2017}. 
It is based on the Shapley value from cooperative game theory \cite{shapley7ValueNPerson1953}, which measures a player's contribution to a team's overall success. 
In the context of ML, the Shapley value is adapted to evaluate the impact of each input feature on the model's prediction.
The SHAP value $\phi_i$ for feature $i$ is computed as
\begin{equation}
    \phi_i = \sum_{S \subseteq F \setminus \{i\}} \frac{|S|!(|F|-|S|-1)!}{|F|!} [f_{S \cup \{i\}}(x_{S \cup \{i\}}) - f_{S}(x_{S})],
\end{equation}
where $S$ represents all subsets of features excluding $i$, and $f_S$ denotes the model's prediction with the subset $S$. 
Calculating the SHAP value thus requires systematically evaluating the model with and without each feature present. 
Performing this calculation for all subsets is computationally expensive ($\mathcal{O}(2^n)$), so approximation methods like ``Kernel SHAP'' and ``Deep SHAP'' are used to speed up the process \cite{lundbergUnifiedApproachInterpreting2017}. 
These methods also provide estimation methods for obtaining the model prediction in the absence of input features the model was trained on, as detailed in \cite{lundbergUnifiedApproachInterpreting2017}.

The total SHAP value for a model prediction is the sum of individual feature contributions, with the overall prediction approximated as
\begin{equation}
    f(x) \approx \sum_{i=1}^{N} \phi_i(x) + \phi_0. 
\end{equation}
Here, $\phi_0$ is the expected model output, i.e., \ the mean prediction across a chosen background dataset.
In particular, a feature's SHAP value quantifies its contribution to driving the model prediction away from the expected or mean prediction.

In our study, features are obtained directly from the skeleton data and categorized into four groups: position ($J$), velocity ($V$), bone ($B$), and acceleration ($A$), further described \cite{tempelAutoGCNgenericHumanActivity2024, tempelLightweightNeuralArchitecture2024}.
The combined SHAP value of a model trained on these features is thus $\phi = \phi_J + \phi_V + \phi_B + \phi_A$. 
The baseline $\phi_0$ is calculated by averaging over $n=100$ randomly sampled data points from the training set $\mathcal{D}_{train}$.
Although SHAP is computationally expensive, approximations allow for relatively efficient model explanations. 
Furthermore, the simple decomposition of the SHAP value into input feature importances provides intuitive explanations of the model. 
SHAP values can also be aggregated across multiple instances to provide global insights into feature importances across a dataset, helping to identify critical elements influencing model behavior.

\subsection{Perturbation}
Perturbation techniques in XAI are used to analyze model sensitivity by altering input data and observing prediction changes \cite{xiongExplainableArtificialIntelligence2024}.
We use the perturbation technique from \cite{tempelExplainingHumanActivity2025} to test the two explanation methods and perturb specific parts of the model architecture related to skeleton body key points.

We use GCNs as the backbone of our architectures, which adapt convolutional neural networks to graph data \cite{yanSpatialTemporalGraph2018}.
With the mentioned advancement over using raw video data as the input, discussed in section \ref{sec:introduction}, GCNs are particularly suitable for skeleton-based HAR. 
In a GCN, nodes aggregate information from neighboring nodes, capturing spatial relationships relevant to human movements.
This involves multiplying the node feature matrix by a normalized adjacency matrix $\mathbf{A}$, identity matrix $\mathbf{I}$, and a learnable edge importance matrix $\mathbf{E}$. 
$\mathbf{A}$ is then split into $\mathbf{A}_j$, such that $\mathbf{A} + \mathbf{I} = \sum_j \mathbf{A}_j$. 
The output is computed as follows \cite{yanSpatialTemporalGraph2018}:

\begin{equation}
    \mathbf{f}_{out} =  \sum_{j} \mathbf{W}_{j} \mathbf{f}_{in} (\mathbf{\Lambda}_{j}^{-\frac{1}{2}} \mathbf{A}_{j} \mathbf{\Lambda}_{j}^{-\frac{1}{2}} \odot \mathbf{E}_{j})
\end{equation}

Here, $\mathbf{f}_{out}$ and $\mathbf{f}_{in}$ are output and input features. 
$\mathbf{W}_j$, $\mathbf{\Lambda}_j$, and $\mathbf{A}_j$ handle convolution and adjacency matrix normalization, while $\mathbf{E}$ (learnable diagonal matrix) has the edge importance for the specific body key point.

As in \cite{tempelExplainingHumanActivity2025}, we apply the perturbation by modifying edges $e_n$ in $\mathbf{E}$ corresponding to the body key points, using the sorted SHAP and Grad-CAM values as the importance measure to choose which edge has to be perturbed.

\section{Experiments}
This section thoroughly describes the experiments conducted on the NTU RGB+D and CP datasets.
First, we explain the two datasets and outline how they are preprocessed for the model.
Then, we give implementation details regarding the GCN model, Grad-CAM, and SHAP within our framework.
Next, we compare how the explainability techniques map importance onto body key points in the skeleton sequences and the overall feature ranking of those. 
Further, we perform the perturbation experiments to investigate which of the most important body key points, as determined by SHAP and Grad-CAM, respectively, lead to a more significant decrease in model performance.
Also, we correlate Grad-CAM values from different layers.
Finally, we compare the computation time of both explainability methods within our experimental setting.

\subsection{Datasets}
\label{sec:dataset}
The NTU RGB+D 60 dataset is a 3D human activity dataset widely used for action recognition tasks \cite{shahroudyNTURGB+DLarge2016}. 
It includes $56,880$ video samples, categorized into 60 distinct action classes. 
We use the skeleton mapping of the videos on $n=19$ body key points for our experiments \cite{shahroudyNTURGB+DLarge2016}.
Our experiments focus on the cross-view (X-View) subset as defined in \cite{shahroudyNTURGB+DLarge2016}. 
This subset divides the dataset based on the recorded camera perspectives. 
Specifically, camera views two and three are used to form the training set, $\mathcal{D}_{train}$, while camera view one serves as the validation set, $\mathcal{D}_{val}$. 
For the scope of this study, we explain three action classes performed by a single individual from $\mathcal{D}_{val}$.

The second dataset consists of 557 videos featuring infants with medical risk factors for cerebral palsy (CP), recorded between 2001 and 2018 across four countries: the United States \((n=248)\), Norway \((n=190)\), Belgium \((n=37)\), and India \((n=82)\).
These videos capture spontaneous movements of infants in a supine position, within the age range of 9 to 18 weeks, and are standardized following the protocols established in \cite{prechtlEarlyMarkerNeurological1997, einspielerPrechtlsMethodQualitative2004}. 
Two medical experts assessed the recordings. 
Each video frame includes the positions of $n=29$ body key points representing the infant's skeleton, recorded as $x$ and $y$ coordinates.
The video sequences undergo several preprocessing steps to ensure consistency and improve data quality. 
These steps include resampling the videos to 30 Hz, applying a median filter to reduce noise, and normalizing the data based on the infant's trunk length, which serves as a reference point. 
The normalization involves doubling the trunk length and centering the data at the median mid-pelvis position. 
After preprocessing, the videos are divided into 5-second windows with a 2.5-second overlap between consecutive windows.
The reader is referred to \cite{groosDevelopmentValidationDeep2022} for more details on this dataset.

\subsection{Implementation}
We utilize two GCN models for the comparison, which is the best-performing model architecture from \cite{tempelAutoGCNgenericHumanActivity2024} for the NTU RGB+D 60 X-View dataset and the architecture from \cite{tempelLightweightNeuralArchitecture2024} for the CP dataset. 
In order to compute SHAP values, we apply the `DeepExplainer'' from the SHAP library \cite{lundbergUnifiedApproachInterpreting2017}.
The ``DeepExplainer'' algorithm requires a reference dataset representing the underlying data distribution.
For this, a random reference set of $n=100$ instances is sampled, covering all action classes generated for the NTU RGB+D dataset. 
Similarly, a reference set consisting of $n=100$ randomly selected window samples from both classes is generated for the CP dataset. 
Since our GPU memory limits the number of background samples that can be processed at once, subsamples of the background ($n=20$) are used in each experiment, with the resulting SHAP values aggregated afterward. 
Given the hardware accessibility of typical medical and research institutions, we recommend this as a practical approach.
The Grad-CAM implementation utilizes the PyTorch \textit{registerhook} functionality at the specified layer to acquire the gradients and feature maps \cite{paszkePyTorchImperativeStyle2019}.
All experiments are conducted on a single NVIDIA V100 GPU with 32 GB of memory, using the PyTorch framework (version 2.3.1) \cite{paszkePyTorchImperativeStyle2019} and shap (version 0.46.0) \cite{lundbergUnifiedApproachInterpreting2017}. 
The global random seed for the experiments is set to $1234$.

\subsection{Qualitative Explanation on Skeleton}
\begin{figure}
    \captionsetup{belowskip=1pt,aboveskip=0pt} 
    \centering
    \begin{subfigure}[b]{0.95\linewidth}
        \centering
        \includegraphics[width=\linewidth]{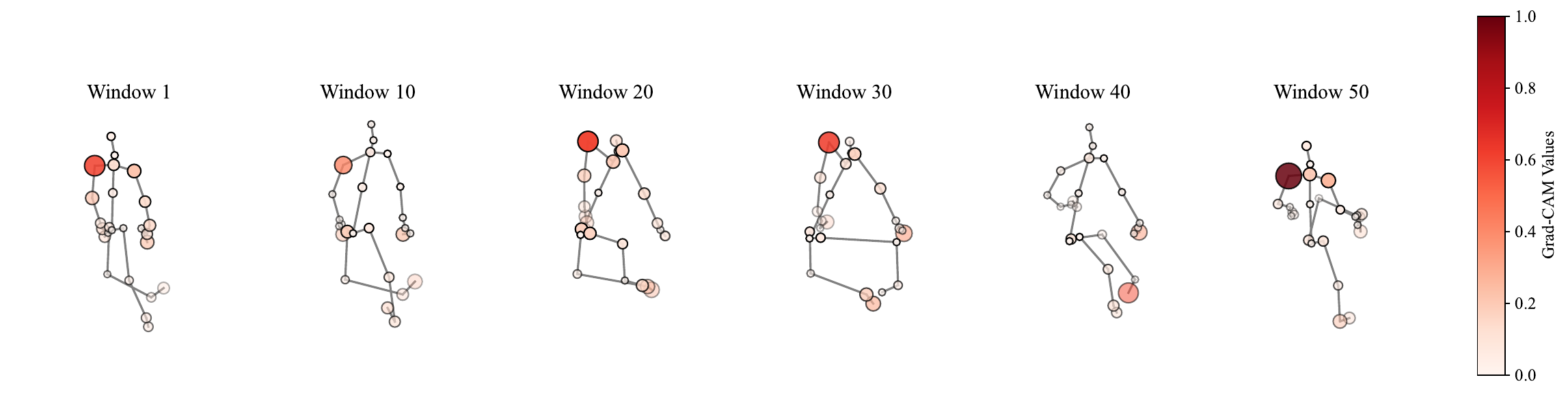}
        \caption{Grad-CAM: Temporal convolutional layer}
        \label{fig:ntu_tcn}
    \end{subfigure}
    
    \begin{subfigure}[b]{0.95\linewidth} 
        \centering
        \includegraphics[width=\linewidth]{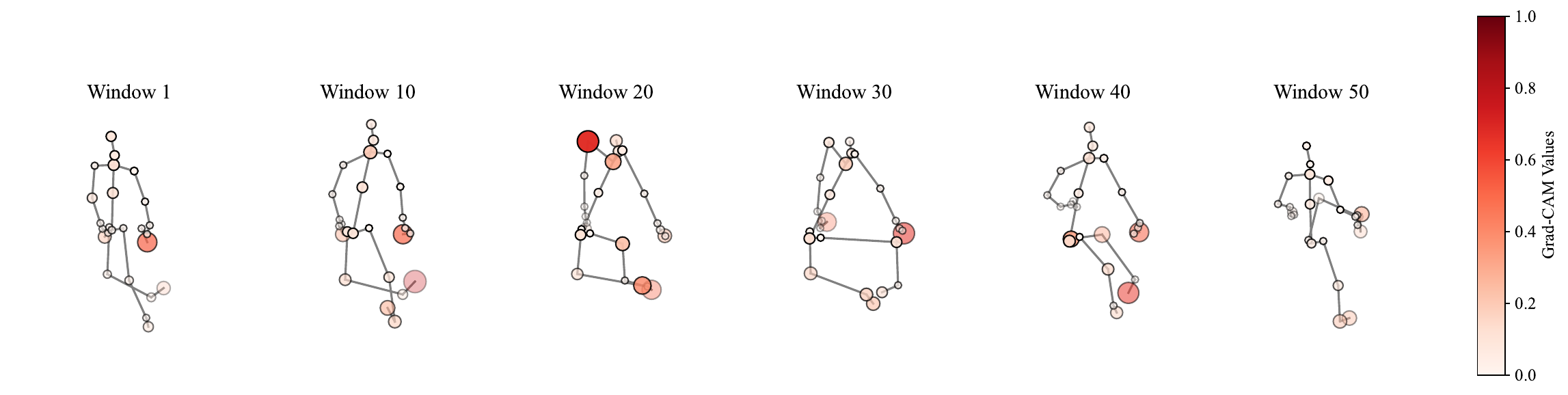}
        \caption{Grad-CAM: Attention activation layer}
        \label{fig:ntu_att}
    \end{subfigure}
    
    \begin{subfigure}[b]{0.96\linewidth} 
        \centering
        \includegraphics[width=\linewidth]{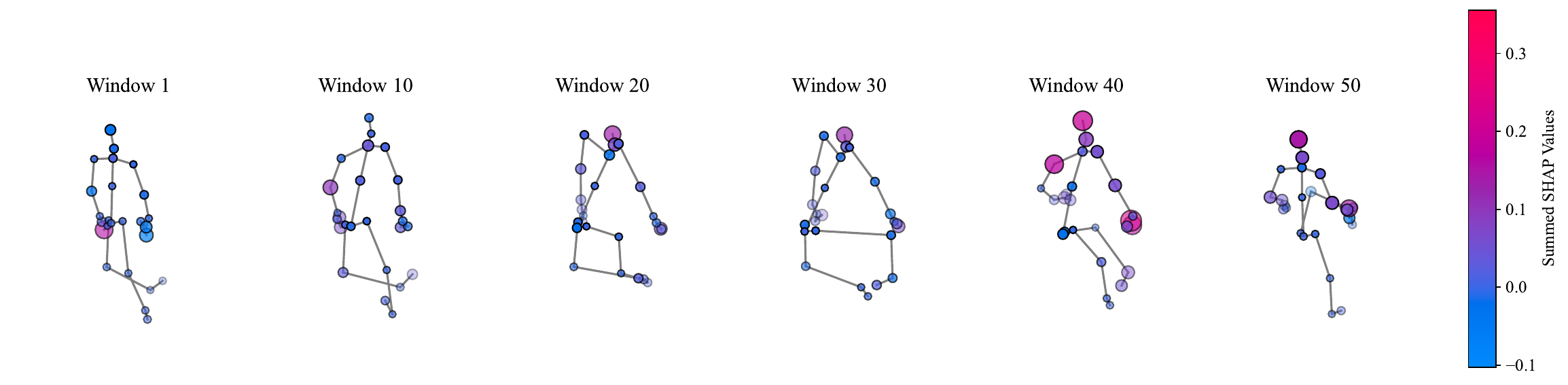}
        \caption{SHAP}
        \label{fig:ntu_shap}
    \end{subfigure}
    
    \caption{Comparison of the spatial explanations on the body key points with Grad-CAM for the TCN convolutional layer and attention activation layer, and SHAP for class 6 (\textit{pick up}) of the NTU RGB+D dataset.}
    \label{fig:ntu_combined}
\end{figure}

\begin{figure}
    \captionsetup{belowskip=1pt,aboveskip=0pt} 
    \centering
    \begin{subfigure}[b]{0.65\linewidth} 
        \centering
        \includegraphics[width=\linewidth]{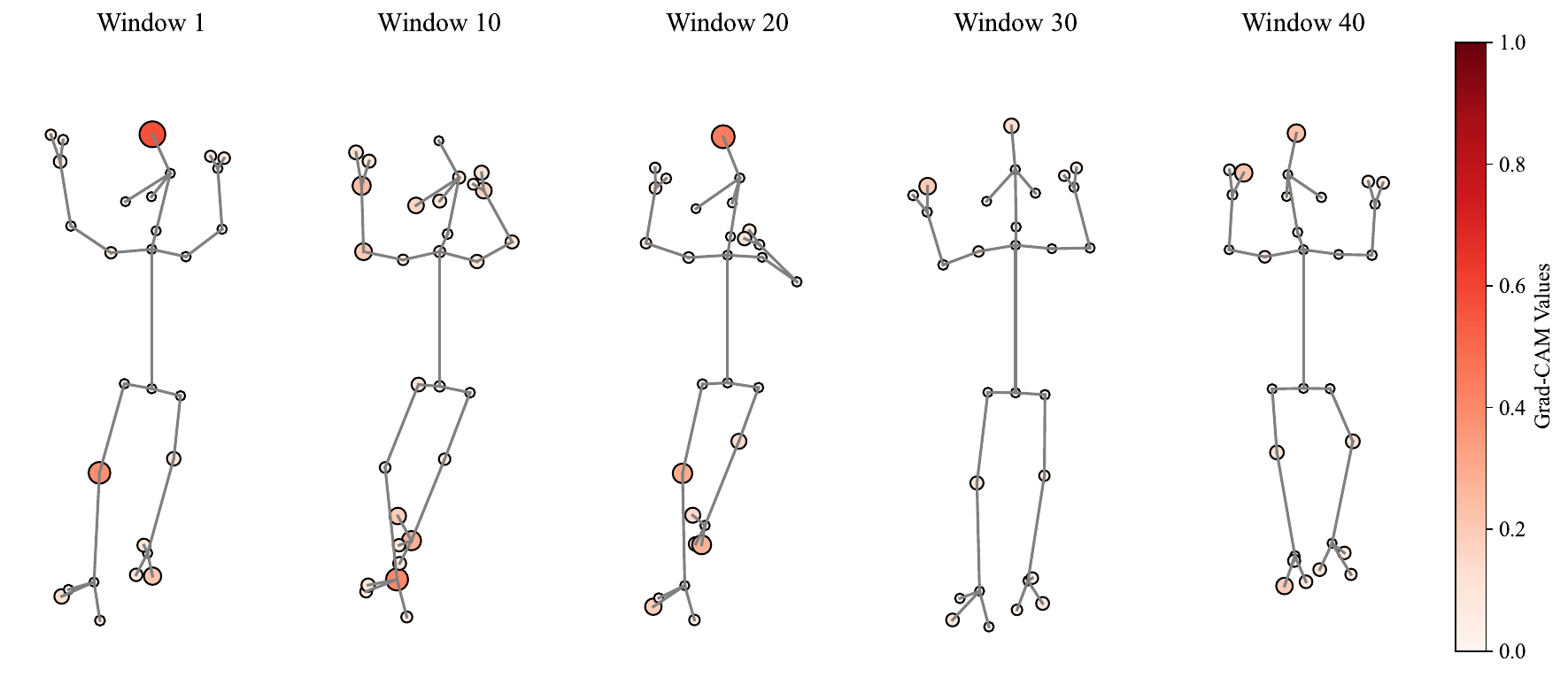}
        \caption{Grad-CAM: Temporal convolutional layer}
        \label{fig:cp_tcn}
    \end{subfigure}
    
    \begin{subfigure}[b]{0.65\linewidth}
        \centering
        \includegraphics[width=\linewidth]{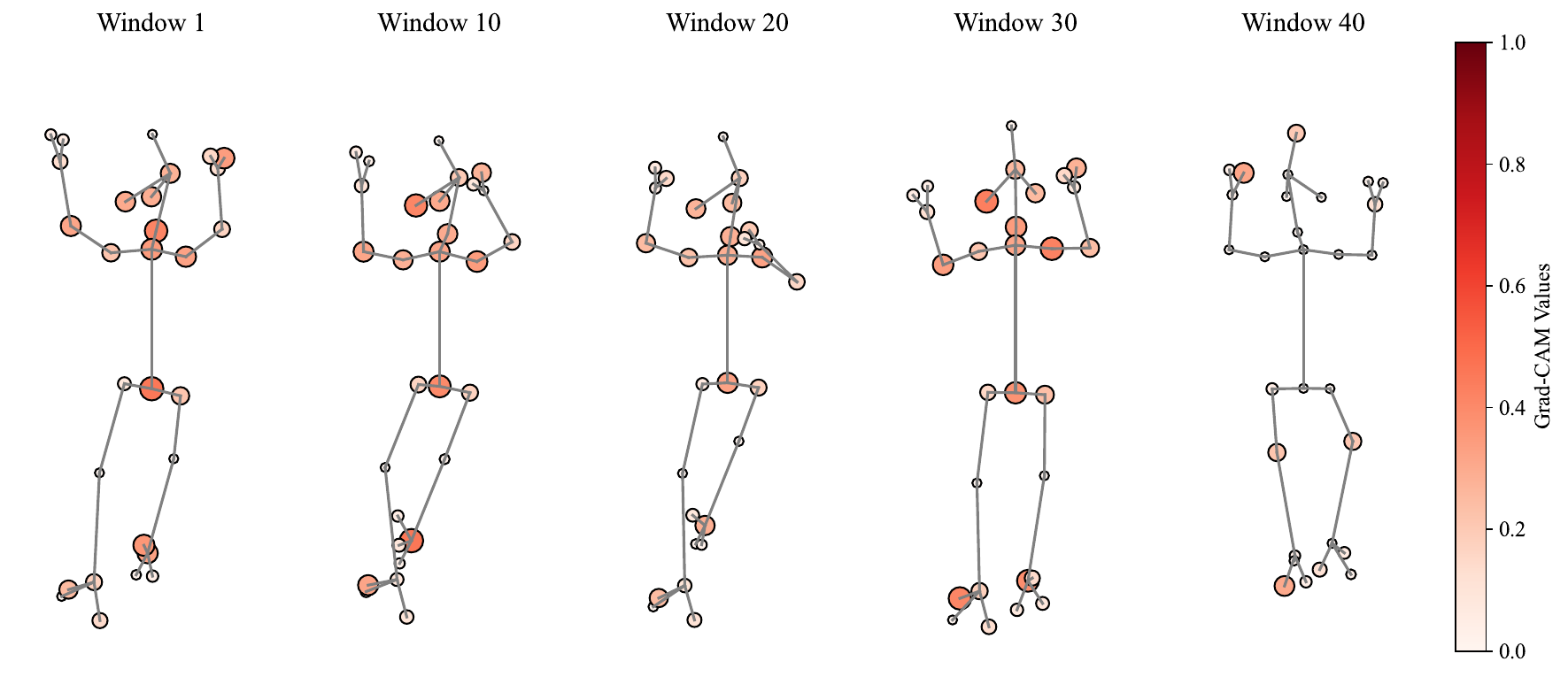}
        \caption{Grad-CAM: Attention activation layer}
        \label{fig:cp_att}
    \end{subfigure}
    
    \begin{subfigure}[b]{0.7\linewidth}
        \centering
        \includegraphics[width=\linewidth]{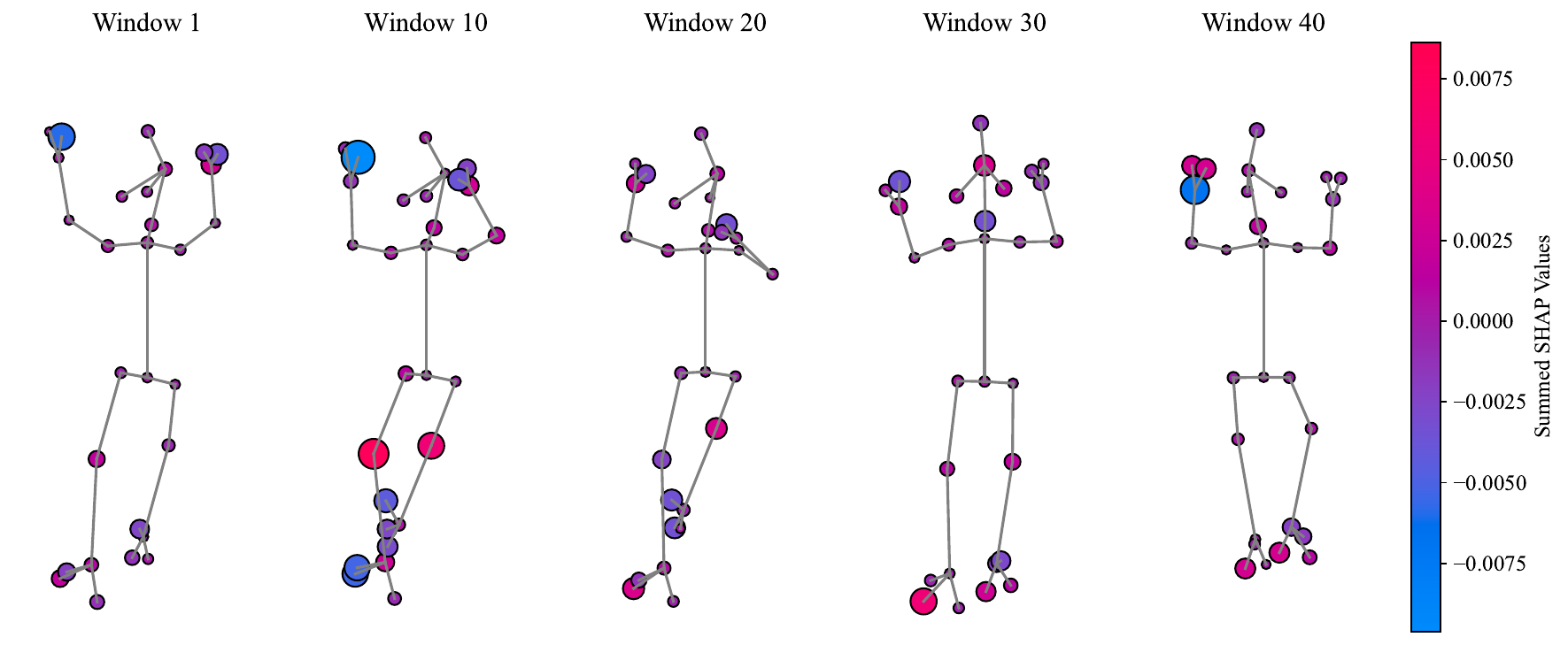}
        \caption{SHAP}
        \label{fig:cp_shap}
    \end{subfigure}
    
    \caption{Comparison of the spatial explanations on the body key points with Grad-CAM for the TCN convolutional and attention activation layers, and SHAP for an infant with CP.
    The size and color indicate the activation in the respective body key point.}
    \label{fig:cp_combined}
\end{figure}

In Fig.~\ref{fig:ntu_combined} and \ref{fig:cp_combined}, the results of the SHAP and the Grad-CAM values are shown when mapped on the skeleton sequence to visualize their contribution at a specific time frame.

For the NTU RGB+D skeleton, which is taken from class 6 (\textit{Pick up}), it can be observed that the Grad-CAM activations for the TCN and attention activation layer differ from each other. 
While the Grad-CAM values for the TCN layer have a higher portion of activated joints, those from the attention activation layer keep a smaller amount of activated joints.
This can be explained by the functionality of the joint attention layer compressing the feature map obtained from the TCN layer to focus on fewer important body key points.
Interestingly, the SHAP and Grad-CAM importance mappings do not agree with each other.
For instance, in window $40$, SHAP attributes higher importance to the upper body, particularly the head and shoulder regions. 
Meanwhile, Grad-CAM emphasizes the left limbs, giving them a higher Grad-CAM value. 
This contrast highlights the differing perspectives of these two XAI techniques on which body regions are most crucial for the model's predictions.

For the CP skeleton analysis, a random infant with CP is selected from the validation set $\mathcal{D}_{val}$, and both Grad-CAM and SHAP values are mapped onto the respective body key points. 
Notably, the two methods do not always align in their activation of body key points. 
This discrepancy is evident in window 40, where each method highlights different body regions. 
However, there is an agreement between SHAP and Grad-CAM activations in the CP dataset compared to the NTU RGB+D dataset, e.g., in window 10, where both methods give the knees a higher activation.
This is potentially due to the smaller window size, which offers more thorough explanations without excessive averaging.

\subsection{Perturbation} \label{sec:perturbation}
\begin{figure}[ht]
    \centering
    \includegraphics[width=1\linewidth]{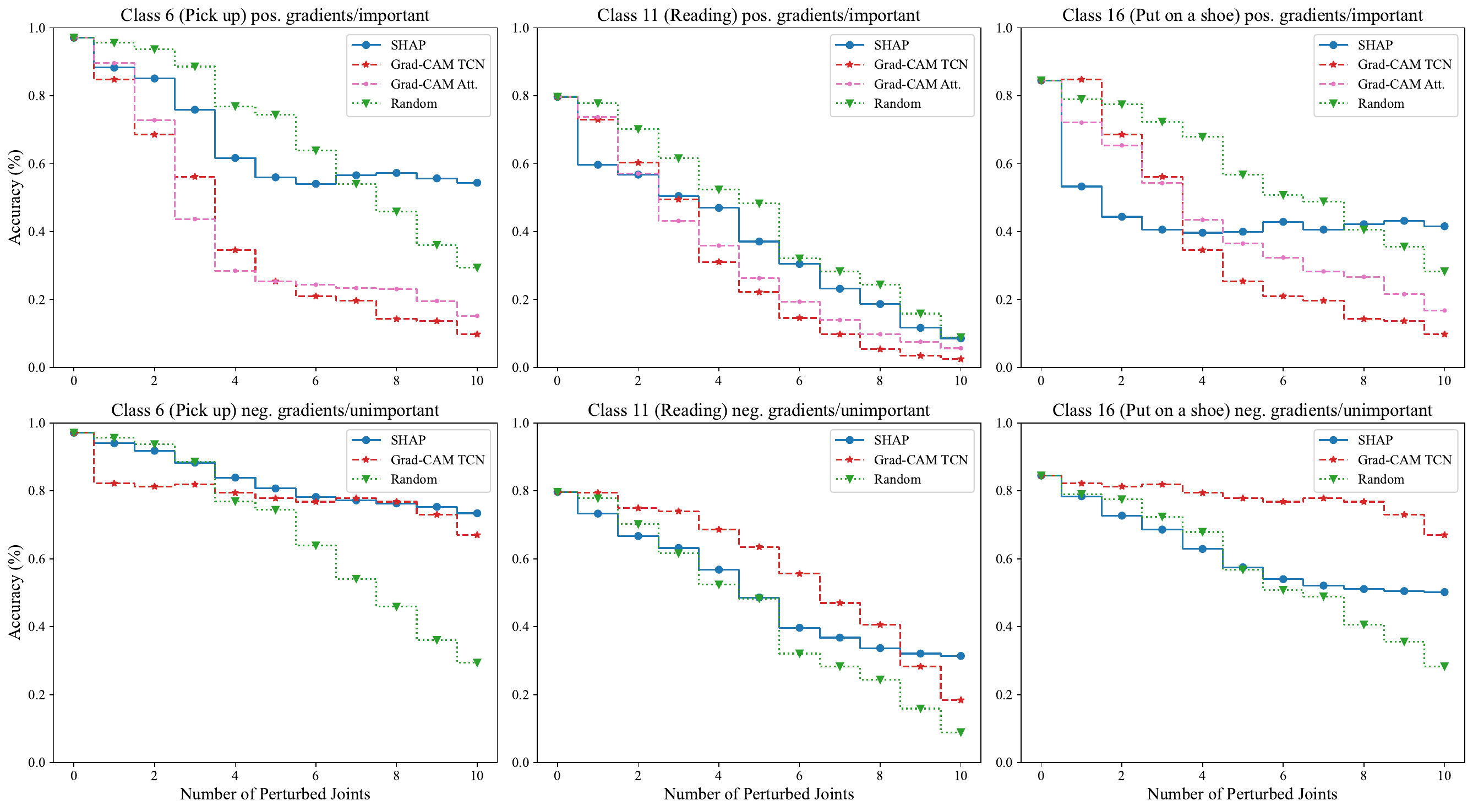}
    \caption{Perturbation experiment results with the three classes from the NTU RGB+D dataset.
    The first row shows the drop in accuracy if important body key points are perturbed.
    The second row shows the results when unimportant body key points are perturbed.
    Both XAI methods perform better than randomly perturbing the body key points, approving their correctness.}
    \label{fig:pert_ntu}
\end{figure}

We apply the same perturbation approach as in \cite{tempelExplainingHumanActivity2025}.
Hence, we can directly compare the influence of perturbing important or unimportant body key points on the CP and NTU RGB+D datasets.
Furthermore, we apply the negated Grad-CAM values on the TCN layer from the model for the NTU RGB+D dataset.
This way, we can also obtain contributions that negatively influence the prediction of the respective class.
For the NTU RGB+D dataset, we chose the same perturbation threshold of $0.35\%$ as in \cite{tempelExplainingHumanActivity2025} to be able to compare the Grad-CAM results against the reported SHAP values from this study.
In Fig.~\ref{fig:pert_ntu}, we show the influence of perturbing up to 10 key body points on the prediction performance for the NTU RGB+D dataset for three classes.

We compare the perturbation results for the NTU RGB+D dataset when taking the TCN and last activation layers' gradients before the classifier against SHAP.
It can be seen that both explanation methods perform better than perturbing edges randomly for important and unimportant features/gradients.
Moreover, perturbing important body key points identified using Grad-CAM outperform those determined by SHAP across all three classes. 
The performance drop is more significant when using the key points identified by Grad-CAM, while SHAP shows a minor degradation when negating gradients or focusing on unimportant features.
While SHAP initially performs on pair or even better than Grad-CAM, the performance decline stops at around six perturbed joints for class 6 (\textit{Pick up}) and four for class 16 (\textit{Put on a shoe}).
For the experiment with negated gradients for Grad-CAM and unimportant features for SHAP - SHAP performs better in class 6 but slightly worse in the two other classes.

\begin{figure}[ht]
    \centering
    \includegraphics[scale=0.7]{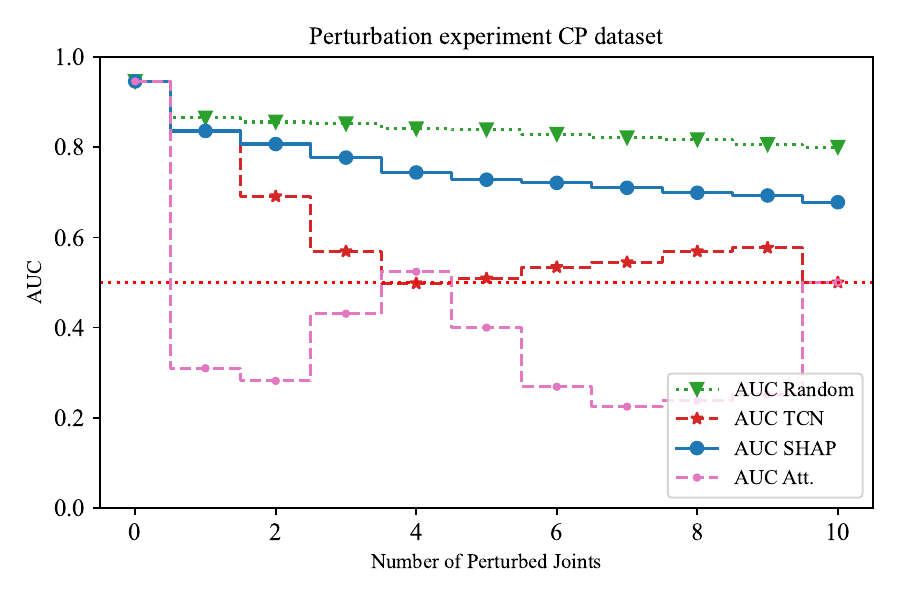}
    \caption{Perturbation results on the CP dataset.
    Both XAI methods perform better than randomly perturbing body key points, indicating their reliability.
    The results when the Grad-CAM values are computed after the frame attention layer perform poorly, indicating that this layer is not well suited for obtaining spatial information.}
    \label{fig:pert_cp}
\end{figure}

We evaluate the perturbation experiment using the Area Under the Curve (AUC) metric for the CP dataset, shown in Fig.~\ref{fig:pert_cp}. 
The results reveal that perturbation of the TCN layer, as assessed through Grad-CAM values, causes a sharp decline in the AUC curve, particularly after perturbing four body key points. 
In contrast, perturbation guided by SHAP explanations leads to a steady AUC decline, implicating a correct determination of the important body key points. 
On the other hand, the perturbation experiments with the Grad-CAM values taken after the attention layer from the main branch produce erratic results. 
This unpredictability can be attributed to the CP model's architecture, where the attention layer primarily focuses on frames and simultaneously activates all joints for an important window. 
Consequently, the perturbation experiments, which depend on correct spatial information of the important body key points, result in uncorrelated AUC oscillation.
In contrast, the joint attention layer for the NTU RGB+D model concentrates on the important body key points rather than time frames, resulting in a smooth performance decrease with the rising number of perturbed edges. 

\subsection{Body key points ranking}
\begin{figure}[ht]
    \centering
    \includegraphics[width=1\linewidth]{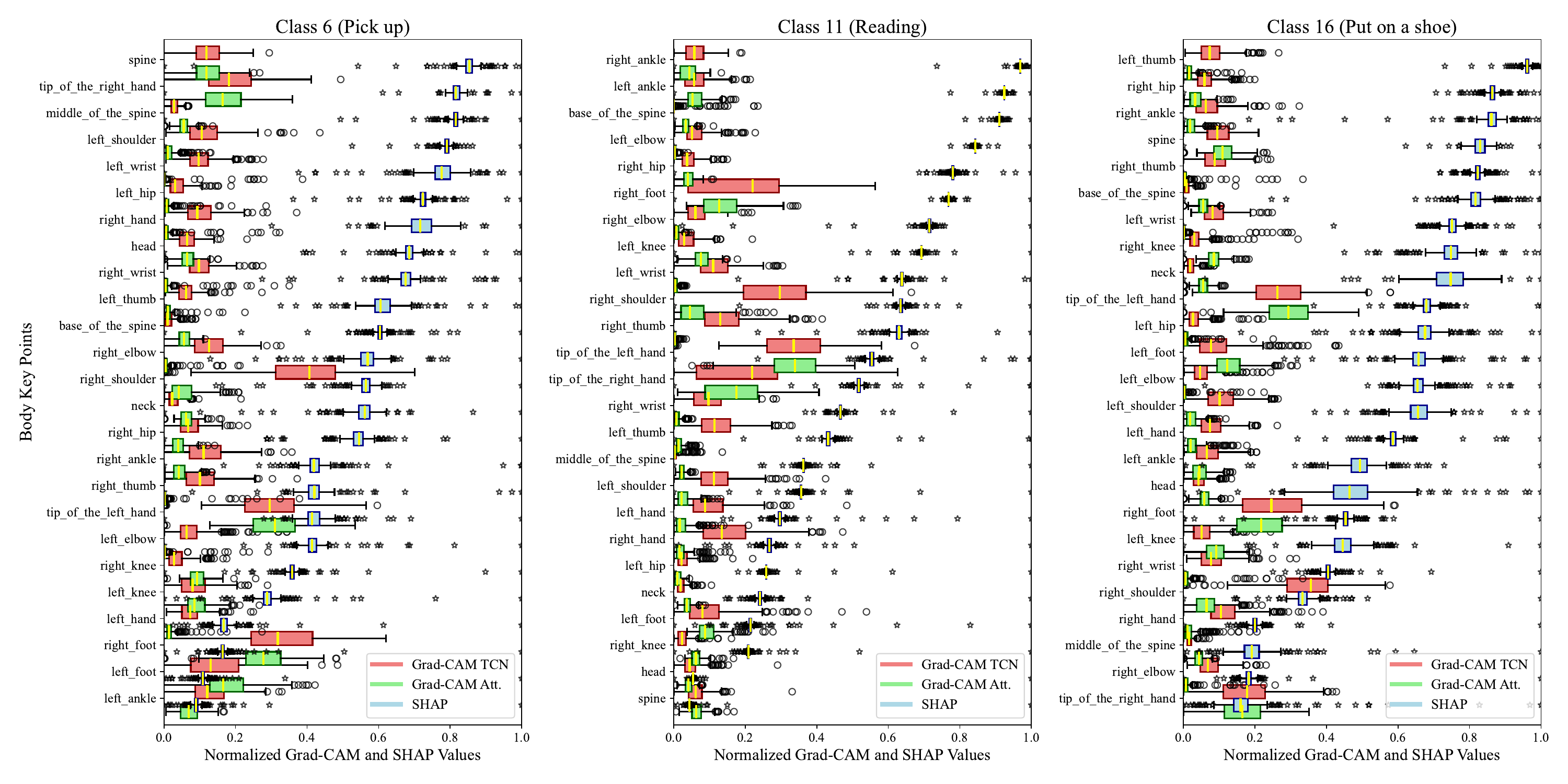}
    \caption{Body key point ranking for the NTU RGB+D dataset on $n=316$ instance of three classes taken from $\mathcal{D}_{val}$. 
    The plot displays the median and interquartile range and highlights outliers for each body key point and explainability method.
    The body key points are sorted based on the median of the SHAP values.}
    \label{fig:ntu_ranking}
\end{figure}

We compare both XAI techniques' body key point ranking for the NTU RGB+D dataset, shown in Fig.~\ref{fig:ntu_ranking}.
It can be seen that SHAP and Grad-CAM disagree on the ordering of the body key points for the NTU RGB+D dataset.
This is likely due to their conceptual difference, where Grad-CAM gives a more spatial-orientated explanation. 
Another potential reason for this differing sorting is the averaging technique, where the Grad-CAM and SHAP values are averaged over the whole sequence.
At the same time, SHAP attributes a more general feature contribution to the interactions with other features for a specific body key point.
While Grad-CAM values have a more extensive spread among the different body key points, SHAP, especially for \textit{Class 11}, has a smaller interquartile range from the median.

\begin{figure}[ht]
    \centering
    \includegraphics[width=1\linewidth]{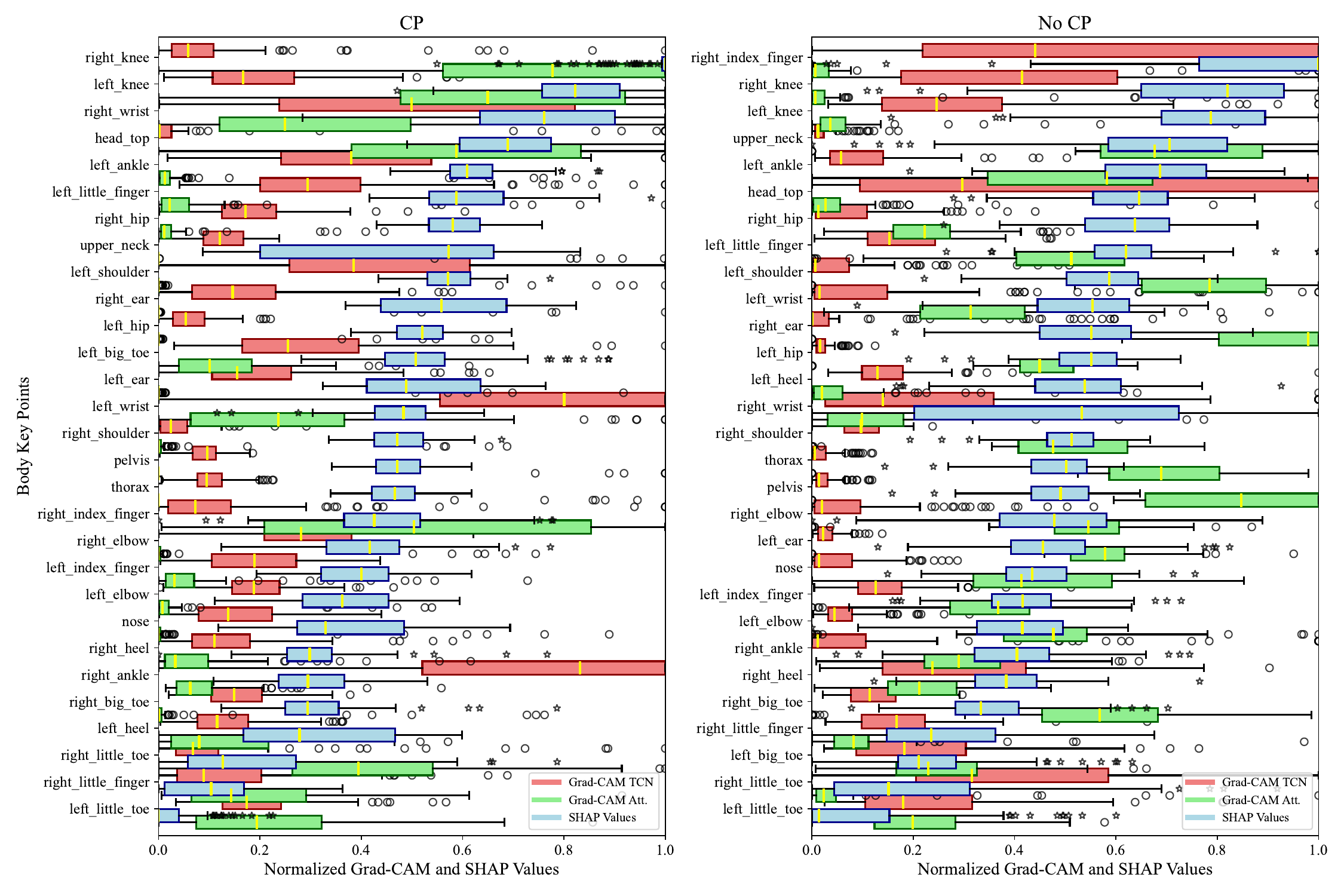}
    \caption{Body key point ranking for one infant with and without CP from $\mathcal{D}_{val}$.
    The plot displays the median and interquartile range and highlights outliers for each body key point and explainability method.
    The body key points are sorted based on the median of the SHAP values.}
    \label{fig:cp_ranking}
\end{figure}

We apply the same body key point ranking for the Grad-CAM and SHAP values to one infant with CP and one without CP, shown in Fig.~\ref{fig:cp_ranking}.
Again, it can be observed that the feature rankings do not agree with SHAP and Grad-CAM values.
Interestingly, the Grad-CAM values from the frame attention and TCN layers do not agree on their body key point ranking. 
The exact reasons, as in Sec.~\ref{sec:perturbation}, can be taken to explain this behavior.
As with the NTU RGB+D dataset, the SHAP values have a smaller interquartile range from the median, indicating a higher certainty for the ranking.
While SHAP gives the highest importance for the infant with \textit{CP} to the \textit{right and left knee} with SHAP, the Grad-CAM values are also the highest in those two body key points for the attention activation layer.
For \textit{No CP}, the most critical features, based on the median value, are the \textit{right index finger} and \textit{right knee}, which also align with the TCN Grad-CAM values.

\subsection{Grad-CAM in Early Layers}
\label{sec:early}

To demonstrate the shortcomings of Grad-CAM in attributing individual input features, we conduct a correlation experiment where Grad-CAM values are computed for different GCN layers, as shown in Figure \ref{fig:pipeline}. 
We assess the correlation of Grad-CAM values in \textbf{1)} initialization layer $L_{\text{in}, i}$, \textbf{2)} TCN layer $L_{\text{TCN}, i}$, and, \textbf{3)} attention activation layer $L_{\text{Att.}, i}$ of input branches $i$ to reference values in the attention activation layer of the main branch.
We also compute correlation for the sum of input branches ($L_{\text{in}}$, $L_{\text{TCN}}$, and $L_{\text{Att.}}$ respectively) and include in the analysis the TCN layer of the main branch $m$ ($L_{\text{TCN},m}$). 

Similar to \cite{dasGradientWeightedClassActivation2022}, layerwise Grad-CAM values are normalized into the range $[0, 1]$ before Spearman's rank correlation coefficient $\rho$ is computed. 
Our results in Table \ref{tab:corr} show that, for the NTU RGB+D dataset and the CP dataset, Grad-CAM values at input branches do not correlate with reference values at the attention activation layer of the main branch. 
The lack of correlation at input branches of the GCN suggests that SHAP is superior to Grad-CAM for attributing individual input features (joint, velocity, bone, and acceleration) and specific input channels.

\begin{table}[ht]
    \centering
    \caption{Correlation of Grad-CAM values obtained from different layers $L$ for the NTU RGB+D and CP datasets.}
    \begin{tabular}{@{}lcc@{}}
        \toprule
        \textbf{Layer}      & \textbf{NTU RGB+D} & \textbf{CP}   \\ 
        \midrule
        $L_{\text{in},j}$   & 0.00 & 0.03 \\
        $L_{\text{in},v}$   & 0.05 & 0.03 \\
        $L_{\text{in},b}$   & -0.01 & 0.02 \\
        $L_{\text{in},a}$   & 0.06 & 0.03 \\
        $L_{\text{in}}$     & -0.03 & 0.04 \\
        $L_{\text{TCN},j}$  & 0.08 & 0.01 \\
        $L_{\text{TCN},v}$  & -0.07 & -0.06 \\
        $L_{\text{TCN},b}$  & -0.04 & 0.01 \\
        $L_{\text{TCN},a}$  & 0.04 & -0.01 \\
        $L_{\text{TCN}}$    & 0.00 & -0.03 \\
        $L_{\text{Att.},j}$ & -0.05 & -0.14 \\
        $L_{\text{Att.},v}$ & -0.06 & 0.11 \\
        $L_{\text{Att.},b}$ & -0.14 & 0.10 \\
        $L_{\text{Att.},a}$ & -0.13 & 0.16 \\
        $L_{\text{Att.}}$   & 0.00 & -0.01 \\
        $L_{\text{TCN},m}$  & 0.24 & -0.13 \\ 
        \bottomrule
    \end{tabular}
    \label{tab:corr}
\end{table}

\subsection{Runtime}
Table \ref{tab:runtime} lists the computation time for Grad-CAM and SHAP within our environment.
We show the time until the individual algorithm arrives at its final explanation.
The Grad-CAM method is computationally faster than the SHAP method for both datasets due to its lower computational complexity and fewer required calculations.
Grad-CAM only necessitates one forward and one backward pass of the obtained gradients.
On the contrary, SHAP requires multiple propagations of the input data on every feature input.
Additionally, the SHAP computation has to be repeated five times in our environment due to memory restrictions to obtain the representative reference dataset of $n=100$.
In contrast, SHAP computes the individual feature contribution for all four input features, while Grad-CAM only provides the gradients of the chosen layer.

\begin{table}[h]
    \centering
    \caption{Runtime for SHAP and Grad-CAM explanation on the CP dataset (both classes) and NTU RGB+D X-View (one class) on $\mathcal{D}_{val}$. 
    The Grad-CAM runtime is for one layer, and the SHAP runtime is for one iteration with the reference dataset $n=20$.}
    \label{tab:runtime}
    \begin{tabular}{@{}lll@{}}
        \toprule
        \textbf{Dataset} & \textbf{Grad-CAM} & \textbf{SHAP}    \\ 
        \midrule
        CP                  & 0.35 hours      & 3.25 hours     \\
        NTU RGB+D           & 0.06 hours      & 3.22 hours     \\ 
        \bottomrule
    \end{tabular}
\end{table}

\section{Discussion}
Our comparison of SHAP and Grad-CAM within HAR demonstrates that both methods fundamentally differ in their methodology and the outcome of the final explanation.
While they both have unique strengths and weaknesses, their applicability depends heavily on specific needs, requirements, and use cases, which we will elaborate on.

Understanding the model's decision-making process is crucial in CP diagnosis or treatment planning. 
SHAP can be used to provide a feature-level explanation, giving clinicians insight into how much each biomechanical feature of each body key point contributes to the model's final decision. 
This type of information is highly valuable in clinical settings where practitioners need clear explanations of the model outputs. 
For example, knowing that certain features like acceleration or velocity are more influential in a diagnosis might help to guide the focus on these particular features.
Grad-CAM provides spatial explanations, which can be helpful in tasks such as reckoning important body key points in the recording at a specific time frame. 
It highlights regions of high activation in the model's chosen target layers, which correspond to body key points in the skeletal structure. 
This explanation is valuable when clinicians need to visualize which body regions are emphasized, though it lacks the granularity of individual feature importance that SHAP provides.

One significant advantage of SHAP is its ability to depict the importance of opposing features. 
With SHAP, it is possible to identify which body key points, in combination with the input features ($J, V, B, A$), are essential for the model's decision and which are unimportant. 
While Grad-CAM can show negative importance by reversing gradients, this remains a spatial explanation and lacks the granularity needed for deeper insights.

Another important consideration is the chosen convolutional layer for the Grad-CAM computation in the model.
Our experiments indicate that the last convolutional layer within the network preserved the most information relevant to the final decision. 
This aligns with prior research recommending focusing on the last convolutional layer for most informative gradient information \cite{selvarajuGradCAMVisualExplanations2020}.
We also showed that the early-layer activations do not correlate with the activations from the TCN or attention layer, which shows the best results in the perturbation experiment.
Furthermore, we obtained the Grad-CAM values from the frame attention layer for the CP dataset, which focuses on the important time frames rather than the body key points.
This may activate body key points, which may no longer provide rich spatial information in some windows.
This is due to the attention mechanism, which focuses on the important frames and activates them as a whole instead of the important body key points.
When using Grad-CAM for the activation of the attention layer, the explanation should be interpreted in relation to a specific type of attention (e.g., temporal frame attention versus spatial joint attention), and the architecture of the GCN itself should be modified to emphasize the most useful explanation for the end-user.

Within our study, we applied the perturbation approach from \cite{tempelExplainingHumanActivity2025}, which may be biased towards the spatial-orientated explanation method: Grad-CAM.
Since Grad-CAM takes the gradients at the specific body key points without needing a reference dataset like SHAP does, it is superior in this specific experimental setting.
Grad-CAM values reflect activation intensities from the chosen layers, which may not be directly comparable to the SHAP feature-based importance values. 
For example, Grad-CAM highlights regions based on body key point activations, whereas SHAP reflects how much each feature contributes to the model's output across the entire reference dataset. 
Additionally, the perturbation method is applied to the whole video frame for the NTU RGB+D data, with the importance scores averaged over the time frame. 
While this approach provides a general overview of the particular action's most crucial body key points, it may obscure critical activations of body key points occurring within shorter temporal windows.
By averaging the whole video, we risk overlooking nuanced movements or short-duration actions, which are essential for understanding complex human activities.
Future work should investigate how a finer temporal averaging technique and localized perturbation strategies can address this limitation. 
For instance, perturbing the model over smaller, temporally-defined video segments could provide more granular insights. 
We might capture more precise and contextually relevant perturbations by classifying shorter video segments and perturbing the network for each segment individually. 
Such a method could yield more accurate explanations of the model's behavior, particularly for brief actions involving subtle and short body movements.

SHAP considers feature interactions and how the importance of a body part can change depending on other body parts, while Grad-CAM does not explicitly consider interactions. 
SHAP provides feature-level attribution, which means it gives importance to the overall contribution of input features and body parts. 
If a body key point is highly correlated with a particular feature in the model, SHAP might attribute it to high importance, regardless of the spatial distribution.
SHAP measures how much the body key point contributes to the model's decision, while Grad-CAM measures how much it contributes to the decision based on its activation. 
Even if both methods show a body key part as important, the contribution might come from a different context:
SHAP values can be influenced by feature interaction and a correlation between the kinematic input features in the model and the underlying reference dataset.
In contrast, Grad-CAM will only consider the direct relationship between the skeleton's spatial features and the decision.
Grad-CAM hereby examines the regions that activate the most in response to the model's prediction.
It focuses on the body key points where the model's decision is most influenced, so the resulting heatmap will highlight specific areas that activate the most. 
However, this might not directly correlate with the feature-level contributions and body key point activations that SHAP captures.

The SHAP explanations depend on the reference dataset \cite{haugBaselinesLocalFeature2021, yuanEmpiricalStudyEffect2023}.
In our approach, the background dataset is randomly sampled from $\mathcal{D}_{train}$ for the two datasets, with $n=100$.
The variance of $\mathbb{E}[f(x)]$ scales with $\frac{1}{\sqrt{n}}$, so a larger reference dataset $\phi_0$ will result in a smaller variance but also results in a more extensive runtime of SHAP \cite{lundbergUnifiedApproachInterpreting2017}.
Recognizing the already significantly longer runtime compared to Grad-CAM, a greater background dataset is not feasible for our architecture and the used dataset.
Yet, future work should incorporate an informed selection of the reference dataset.
This can include choosing videos from the CP dataset, which have different severities of CP and are either positive or negative. 
Finally, the SHAP values combined with the perturbation experiment should be investigated to determine if they lead to the same reliability outcomes and to which extent the explanations match the randomly chosen reference dataset.

Runtime and computation needs are critical when considering these methods for clinical use. 
Grad-CAM's faster runtime makes it suitable for real-time analysis or applications where speed is essential.
SHAP, while slower, provides richer information, making it better suited for offline analysis or in-depth clinical studies where speed and computing resources are less of an issue.

Given these methods' different strengths and weaknesses, some recommendations can be made for their application in healthcare environments. 
Grad-CAM might be preferred when a quick, spatial understanding of which body parts the model focuses on is sufficient. 
This could be useful for real-time diagnostic tools or when working with visual data, such as motion capture or video-based assessments. 
However, Grad-CAM may miss the ``bigger picture'', mainly when interactions between body key points and their features are crucial to the diagnosis and explanation.

SHAP inherently satisfies a set of axioms: local accuracy, missingness, and consistency, which provide theoretical guarantees for its explanations based on the Shapley value \cite{lundbergUnifiedApproachInterpreting2017, shapley7ValueNPerson1953}.
These axioms ensure that SHAP reliably distributes the model's output among the input features, excludes non-contributing features, and maintains consistency in attribution when feature contributions change. 
In \cite{molnarInterpretableMachineLearning2022}, the author claims that SHAP might be the only explanation method that satisfies all necessary axioms for reliable model explanation in healthcare.
Consequently, these foundational properties make SHAP a more robust and theoretically sound explanation method compared to Grad-CAM, which does not include such axioms.

While SHAP provides detailed, feature-level explanations, its effectiveness in capturing the true importance of features may be impacted by feature dependencies \cite{aasExplainingIndividualPredictions2021}. 
SHAP's default independence assumption may lead to diluted feature importance in scenarios with correlated features, such as the kinematic chain in human motion.
Grad-CAM, with its spatial focus and faster runtime, is more suited to applications where visual feedback and quick assessments are needed. 
The two methods can complement each other, with SHAP giving the ``why'' behind model decisions and Grad-CAM giving the ``where''.
In conclusion, the explainability of an ML model should always be tailored to different stakeholders, providing each with an appropriate level of detail based on their specific needs \cite{koricaExplainableArtificialIntelligence2021}.

\section{Limitations and Future Work}
While our approach of comparing Grad-CAM with SHAP shows the individual strengths and weaknesses, limitations exist, which open possibilities for future work.

Our comparison study is restricted to the two explanation methods, SHAP and Grad-CAM, which are the typical approaches applied for GCNs within the HAR domain.
While these two techniques are the common choice when explaining skeleton data, other explanation methods should be discovered.
This can include the predecessor of Grad-CAM, Grad-CAM++ \cite{chattopadhyayGradCAMImprovedVisual2018}, a generalized approach that improves localization capability.
In particular, Grad-CAM++ may produce more localized explanations, which could be validated with the perturbation experiment.
Another possible explanation method that could be used is LIME \cite{ribeiroWhyShouldTrust2016}.
However, perturbing skeleton data is challenging since naive alterations of the sequence can produce unrealistic or biomechanically implausible actions and movements. 
One potential solution is to define kinematic sound perturbations that respect natural joint movement constraints by simulating only plausible variations or applying domain-specific transformations. 
This highlights a key issue in applying LIME to skeleton-based HAR: ensuring that the perturbed data remains realistic so that the local surrogate model returns valid insights.
Future work should, therefore, incorporate the mentioned explanation methods to provide the bigger picture beyond the two applied XAI methods within this work.

In our current work, we apply the two quantitative metrics, PGI and PGU, to assess the reliability of the obtained explanations \cite{agarwalOpenXAITransparentEvaluation2024}, acknowledging that there exists more metrics to assess the reliability of an explanation.
The PGI and PGU metrics are particularly useful in measuring the faithfulness of the explanation when no ground truth is available, which is the case within our approach.
However, future work should explore metrics to measure the explanation stability, such as Relative Input Stability (RIS), Relative Representation Stability (RRS), and Relative Output Stability (ROS), to measure the explanation change when perturbing parts of the model parameters \cite{agarwalRethinkingStabilityAttributionbased2022}.
With these additional metrics, the stability of the explanation can be assessed, contributing to a multifaceted view of the reliability of the obtained explanations.

Another promising avenue for future research is the incorporation of human-centric explanations, which are interrogative. 
In \cite{sporsemCliniciansDontKnow2025}, the authors show that clinicians prefer such interrogative explanations, where they can explore the output of the explanations on their own.
By providing multiple explanation methods, clinicians can choose the most suited one for their specific use case.
Therefore, it is essential to communicate the individual distinctions of each explanation method, such as the spatial orientation of Grad-CAM or the granular explanation capabilities of SHAP, for which this comparative study lays the groundwork.

Depending on the specific demands of the task, whether it is the need for quick, spatial understanding or a thorough, feature-level explanation, these methods can be used in parallel to provide a complete explanation of the model behavior on the dataset.
For future work, hybrid approaches that combine the strengths of both SHAP and Grad-CAM should be explored.
This could be achieved through a combined XAI score leveraging the Grad-CAM for spatial explanation and SHAP for feature-level explanation, resulting in an overall score that balances spatial and feature-level explanations.


\section{Conclusion}
This paper systematically compared SHAP and Grad-CAM within the context of GCNs and HAR using skeleton data from two real-world datasets. 
Our findings highlight that these two explainability methods differ in their approach and output, each offering different benefits depending on the specific requirements of the individual use case.

SHAP provides a detailed, feature-level explanation, making it practical in scenarios where it is crucial to understand each individual body's key points' contribution in combination with the input features. 
It offers a more profound, finer view of how specific features impact the model's decisions, which is essential for granular insights in sensitive domains. 
This may be required in healthcare applications such as CP prediction, where model decisions must be transparent and explainable to clinicians. 
However, its computational intensity and longer runtime make it less suitable for real-time applications.

In contrast, Grad-CAM provides quicker, spatially oriented explanations, showing which regions of the body are most influential in the model's decision at a specific time frame. 
Its speed and simplicity make it a better choice for real-time assessments or scenarios where quick and more straightforward visual feedback is necessary.
However, its lack of granularity and inability to capture complex feature interactions from earlier layers limits its effectiveness in circumstances demanding granular feature-level insights.

Finally, our study demonstrates that SHAP and Grad-CAM should be viewed as complementary tools rather than competing methods. 
SHAP provides the ``why'' behind decisions, while Grad-CAM gives the ``where''.


\section*{Contributions}
\textbf{Felix Tempel:} Conceptualization; Methodology; Formal analysis; Software; Writing - original draft preparation.
\textbf{Daniel Groos:} Formal analysis; Methodology; Writing Sec.~\ref{sec:early} - review and editing.
\textbf{Espen Alexander F. Ihlen:} Data curation; Supervision; Writing - review and editing.
\textbf{Lars Adde:} Conceptualization; Data curation; Writing - review and editing.
\textbf{Inga Strümke:} Project administration; Supervision; Writing - review and editing.

\section*{Data Availability}
The NTU RGB + D 60 dataset supporting the findings of this study is available at \url{https://rose1.ntu.edu.sg/dataset/actionRecognition/} \cite{shahroudyNTURGB+DLarge2016}.
Due to ethical considerations, the CP dataset is not publicly available.

\section*{Code availability}
The code and the experimental results are publicly available at \url{https://github.com/DeepInMotion/ShapGCN}

\section*{Acknowledgements}
This work was partly supported by the PERSEUS Project, a European Union’s Horizon 2020 Research and Innovation Program under the Marie Skłodowska-Curie under Grant 101034240, and partly by the Research Council of Norway under Grant 327146.

\printbibliography


\end{document}